\documentclass[nohyperref]{article}

\usepackage{microtype}
\usepackage{graphicx}
\usepackage{subfigure}
\usepackage{booktabs} 
\usepackage{multirow}
\usepackage{hyperref}

\usepackage[accepted]{icml2022}

\usepackage{amsmath}
\usepackage{amssymb}
\usepackage{mathtools}
\usepackage{amsthm}

\usepackage[capitalize,noabbrev]{cleveref}

\theoremstyle{plain}

\theoremstyle{definition}

\theoremstyle{remark}

\usepackage[textsize=tiny]{todonotes}

\icmltitlerunning{Flexible Triggering Kernels for Hawkes Process Modeling}
\begin{document}

\twocolumn[
\icmltitle{Flexible Triggering Kernels for Hawkes Process Modeling}




\begin{icmlauthorlist}
\icmlauthor{Yamac Alican Isik}{xxx}
\icmlauthor{Connor Davis}{xxx}
\icmlauthor{Paidamoyo Chapfuwa}{comp}
\icmlauthor{Ricardo Henao }{xxx}
\end{icmlauthorlist}

\icmlaffiliation{xxx}{Department of Biostatistics and Bioinformatics, Duke University, Durham, USA}
\icmlaffiliation{comp}{Department of Health Policy, Stanford University, Stanford, USA }

\icmlcorrespondingauthor{Yamac Alican Isik}{yamac.isik@duke.edu}

\icmlkeywords{Machine Learning, ICML}
\vskip 0.3in
]


%
\printAffiliationsAndNotice{}  

\begin{abstract}
Recently proposed encoder-decoder structures for modeling Hawkes processes use transformer-inspired architectures, which encode the history of events via embeddings and self-attention mechanisms.
These models deliver better prediction and goodness-of-fit than their RNN-based counterparts.
However, they often require high computational and memory complexity requirements and sometimes fail to adequately capture the triggering function of the underlying process.
So motivated, we introduce an efficient and general encoding of the historical event sequence by replacing the complex (multilayered) attention structures with triggering kernels of the observed data.
Noting the similarity between the triggering kernels of a point process and the attention scores, we use a triggering kernel to replace the weights used to build history representations.
Our estimate for the triggering function is equipped with a sigmoid gating mechanism that captures local-in-time triggering effects that are otherwise challenging with standard decaying-over-time kernels. 
Further, taking both event type representations and temporal embeddings as inputs, the model learns the underlying triggering type-time kernel parameters given pairs of event types.
We present experiments on synthetic and real data sets widely used by competing models, while further including a COVID-19 dataset to illustrate a scenario where longitudinal covariates are available.
Results show the proposed model outperforms existing approaches while being more efficient in terms of computational complexity and yielding interpretable results via direct application of the newly introduced kernel.
\end{abstract}

\section{Introduction}
\label{introduction}
%
%
Temporal point processes are the preferred choice when modeling asynchronous (irregularly sampled) event sequences \citep{cox1980point}.
Of particular interest is the Hawkes Process \cite{10.2307/2334319}, a self-exciting point process that has become popular and is used in numerous applications.
In finance, the Hawkes process is used to model market return events, volatility, and market stability  \cite{doi:10.1080/14697688.2017.1403156,LEE2017174,bacry2015hawkes}.
In social media, it is employed to analyze malicious activity, and actions of users on social media platforms  \cite{8855306,rizoiu2017tutorial}.
More recently, Hawkes processes have become critical in healthcare settings, such as modeling adverse drug reactions, disease progression, and spread of COVID-19 outbreaks \cite{pmlr-v68-bao17a,CHIANG2021,9502579} %
%

\citet{10.2307/2334319} first introduced the Hawkes process to model the aftershocks of earthquakes.
The self-excitation mechanism allowed for modeling the clustered behavior of aftershock occurrences. \citet{ozaki1979maximum} further introduced the likelihood function for the conditional intensity for the parametric estimation of the intensity function.
Non-parametric approaches based on the expectation-maximization (EM) algorithm have also been proposed to estimate the conditional intensity function \cite{lewis2011nonparametric,chen2016nonparametric}.
However, these approaches suffer from expensive computational budgets needed to solve additional ordinary differential equation terms or the choice of the kernel functions needed to estimate the intensity function.

More recently, deep learning approaches have been introduced to model the event sequences with recurrent neural networks (RNNs) and attention-based architectures.
\citet{du2016recurrent} and \citet{mei2016neural} model the intensity as a non-linear function of the sequence history by using RNNs to characterize the behavior of past events and feed-forward decoders to model the intensity function.
These approaches aim to provide a flexible estimation of the conditional intensity function by avoiding strict assumptions on its functional form.
However, the lack of a closed-form solution for the intensity function necessitates computationally expensive approximation techniques such as Monte Carlo to calculate the log-likelihood function values.
\citet{shchur2019intensity} propose to replace the decoder used in the previous models with a log-normal mixture structure.
By focusing on the conditional density function for the arrival times, they produce a closed-form solution for the expectation of the process.
Their approach circumvents expensive approximations and yields efficient sampling of new sequences at the expense of assuming that all arrival times follow a log-normal distribution.
Moreover, these models inherit the issues associated with RNNs, such as vanishing or exploding gradients \citep{pmlr-v28-pascanu13}, or the inability to capture long-term or very short-term dependencies \citep{pmlr-v119-zuo20a}.

Attention-based models outperform their RNN counterparts because of their ability to capture dependencies irrespective of context length.
Models with architectures similar to the transformer \citep{46201} encode the historical influence of past events with multiple attention layers and multiple attention heads.
\citet{pmlr-v119-zhang20q}, and \citet{pmlr-v119-zuo20a} introduced transformer-based approaches for Hawkes process modeling.
These models create history embeddings based on a scoring function that compares the similarity of input vectors for each event in representation space.
Though these attention-based structures can predict the arrival times relatively well, they suffer from high computational and memory requirements.
Their results can be more interpretable compared to RNN-based approaches, but interpretability is limited to the relative contribution of past events.
More importantly, these attention-based approaches struggle to infer the underlying intensity distribution of the point processes.
The predefined attention score function, {\em e.g.}, the exponentiated inner product, limits their ability to capture the triggering kernels of the point process. 
Besides, these models fail to differentiate between triggering kernels for various event type combinations, and do not consider (longitudinal) covariates into their modeling strategy.

Motivated to address the challenges faced by RNN- and attention-based approaches to the modeling of Hawkes processes, we propose a method that replaces the complex attention mechanisms with a more efficient, flexible, and interpretable encoding structure. 
Specifically, our approach replaces the attention mechanism with the direct estimation of the triggering kernels by taking advantage of the resemblance between the attention scores used to create the history embedding and the triggering function of the Hawkes process itself.
The resulting encoder is more efficient than both RNNs and multi-layered attention structures, while being able to learn the triggering kernels of the underlying distribution directly from data.
The proposed kernel uniquely combines a sigmoid gate \cite{Steinruecken2019} with the rational quadratic function enabling the learning of a more general family of triggering kernels, such as local-in-time triggering, as well as the standard decaying-over-time functions.
To our knowledge, this is the first approach that experiments with different structures to produce general triggering kernels geared toward point process modeling.

In addition to flexibility, the proposed {\em sigmoid gated Hawkes process} (SGHP) model can also learn a different triggering kernel for each pair of event types.
Utilizing the event type embeddings previously used by existing attention-based models, we can readily estimate a triggering kernel for each event type combination while preventing the rapid growth of parameters, which typically scales with the square of the number of event types.
Estimating kernels for each event type pairing allows one to interpret the effect of observing a given event type conditioned on other event types by examining their intensity function.

In order to showcase the flexibility, efficiency, and interpretability of the proposed model, we present experiments on both artificial and a wide range of real-world data; including comparisons with the state-of-art models.
Results indicate that our model outperforms all the baseline models in terms of arrival time prediction and reproducing underlying triggering kernels, while being very competitive for event type prediction.
Finally, we present results on a timely COVID-19 dataset of the patients admitted to the emergency department at [Name] Hospital and highlight the interpretability of our model via the learned triggering kernels for several event type pairings.

\section{Background}
%
Let $S$ be a sequence consisting of $L$ observed events (timestamps) from $K$ different possible event types.
Each sequence $S$ can be represented as $S_n = \{(u_i,t_i)\}_{i=1}^{L_n}$, where $u_i \in \{1,2, ... K\}$, and $u_i$ and $t_i$ correspond to the type and timestamp of an observed event, respectively.
We use the subscript $n$ to represent an individual sequence, and $i$ and $j$ to denote events within a given sequence.
Our objective is to model the underlying distribution of the collection of sequences for a given dataset $\{S_n\}_{n=1}^N$ of $N$ sequences, as well as predicting the {\em arrival times} and {\em event types} of future events conditioned on the {\em history} of past events.

\subsection{Temporal Point Process}
Temporal point processes are most often used to model the distribution of event sequences.
Specifically, each temporal point process is characterized by its unique conditional intensity function $\lambda(\cdot)$, defined as
\vspace{-4mm}
\begin{equation}
    \lambda(t|h_t) = \frac{f(t|h_t)}{1-F(t|h_t)} ,
\end{equation}
where $h_t$ is the history of previous events for $t_i<t$, {\em i.e.}, $h_t=\{(t_1,k_1),(t_2,k_2),\ldots,(t_i,k_i)\}$, $f(t|h_t)$ is the conditional probability density function for the (next) event at time $t$ given the history of the sequence and $F(t|h_t)$ is the corresponding conditional cumulative distribution function.

\subsection{Hawkes Process}
The Hawkes process \citep{10.2307/2334319} is a special case of a temporal point process in which the intensity function is defined as follows
\begin{equation}\label{eq:lamba_uni}
    \lambda(t) = \mu(t) +\sum_{i \in h_t}\phi(t-t_i) ,
\end{equation}
where $\mu(t)$ is the exogenous background intensity and $\phi(\cdot)$ is the {\em triggering kernel} that allows the intensity function to depend on past events.
Note that \eqref{eq:lamba_uni} is fully additive provided that both $\mu(t)$ and $\phi(\cdot)$ are non-negative functions.
Moreover, in some practical scenarios, the background intensity is assumed to be constant \citep{Rasmussen2009GaussianPF}.
Similarly, for the multivariate (multi-event) case
\begin{equation}\label{eq:lamba_multi}
    \lambda_k(t) = \mu_k(t) +\sum_{i \in h_t}\phi_{k_{i}k}(t-t_i) , 
\end{equation}
where $\phi_{k_{i}k}(\cdot)$ represents the triggering kernel of an event of type $k_i$ on the current event type $k$, and $\mu_k(t)$ is the background intensity for type $k$.

\subsection{History Representations via Self Attention}
Models based on self-attention have been proposed to capture the pairwise influence of all previous events in a sequence as a means to create historical representations \citep{pmlr-v119-zhang20q,pmlr-v119-zuo20a}.
Specifically, the historical summary for event $j$ in the sequence is represented as follows
\begin{equation}\label{eq:att}
    h_j = \Bigg( \sum_{i < j} f(x_j,x_i)x_i \Bigg) /  \sum_{i < j} f(x_j,x_i) ,
\end{equation}
where $x_j$ is the latent vector (embedding) of the $j$-th event in the sequence and is defined below, and $f(\cdot,\cdot)$ is a similarity function usually specified as the exponentiated inner product between two embeddings, {\em i.e.},
\begin{equation}\label{eq:sim}
    f(x_j,x_i) = \exp(x_j,x_i^T) .
\end{equation}
Note that other similarity functions are readily available and can be used in practice. However, \eqref{eq:sim} is still probably the most popular default choice.

\subsection{Temporal Encoding}
Attention-based models do not intrinsically capture (positional) information about the order of the events in a sequence like, for instance, RNN-based models do through recurrence.
\citet{pmlr-v119-zuo20a} introduced positional embeddings to account for such information into their encoding architecture.
Further, \citet{pmlr-v119-zhang20q} modified these embeddings for event sequences by incorporating event timestamps into the positional embeddings.

These modified positional embeddings, also know as {\em temporal embeddings}, are represented as $t_i^{\rm enc}$ for the $i$-th event occurring at time $t_i$.
Assuming a temporal embedding in $D$ dimensions, each of its elements, $d$, are defined as follows
\begin{align}\label{eq:time_emb}
    t^{\rm enc}_{i,d} = 
                    \begin{cases}
                        \sin( w_d  i +\omega_d t_i), & \text{if $d$ is even} \\
                        \cos( w_d i +\omega_d t_i) , & \text{if $d$ is odd}
                    \end{cases},
\end{align}
where $w_d$ is defined as $w_d=1/(10000^{2d/D})$.
%
%
Conceptually, $w_d$ represents the angular frequency of the $d$-th dimension and
$\omega_d$ is a scaling parameter that controls the weight of the time-shift and is learned from the data.

\subsection{Event Type Embedding}
To represent each event type in the sequence, a linear embedding layer is specified as in \citet{pmlr-v119-zhang20q} via
\begin{equation}\label{eq:type_emb}
    e_{i}^{\rm enc}=e_{i}W ,
\end{equation}
where $e_{i}^{\rm enc}$ is a $D$-dimensional (dense) vector representing the embedding for event at time $t_i$, $e_i$ its one-hot encoding, and $W \in \mathbb{R}^{KxD}$ is the embedding matrix, and is also learned from data.
Note that there are only $K$ distinct $e_{i}^{\rm enc}$ event type embeddings, which (in a slight abuse of notation) are indexed in \eqref{eq:type_emb} over time using the subscript $i$.

\section{Replacing Attention with a Kernel}
Below we describe the proposed {\em sigmoid gated kernel} and explain the way in which we use it as a replacement for the attention mechanism in existing architectures \citep{pmlr-v119-zuo20a,pmlr-v119-zhang20q}.

\subsection{Self Attention via the Triggering Function}
We start by noting the similarity between the weights of the previous events for the attention mechanism in \eqref{eq:lamba_uni} obtained via $f(\cdot,\cdot)$, and the the triggering kernel $\phi_{k_ik}(\cdot)$ in \eqref{eq:lamba_multi}, for a conventional Hawkes process.
Note that the triggering kernel $\phi_{k_ik}(\cdot)$ captures the extent of the influence of past events, for $t_i<t$, on the (future) event at time $t$. 
Similarly, $f(x_j,x_i)$ captures the influence of a past event $x_i$ on the future event $x_j$, by proxy, leveraging representations $x_i$ and $x_j$, for which we have $x_i=t_i^{\rm enc}+e_i^{\rm enc}$, that encodes both temporal and event type information \cite{pmlr-v119-zhang20q}.
A significant difference is that in the standard Hawkes process, a triggering kernel is specified for each event-type pairwise combination, whereas in models based on the self-attention mechanism, such information is encapsulated in the event type embedding via \eqref{eq:type_emb}.
Alternatively, we propose to directly learn a flexible triggering kernel combinations for event types while still considering latent representations for the events in the history of events.

We now define the historical representation for the $j$-th event in the sequence as follows
\begin{equation}\label{eq:h_q}
    h_j = \sum_{i\leq j}q_{k_i k_j}(|t_i - t_j|)x_i ,
\end{equation}
%
where $q_{k_ik_j}(\cdot)$ is a triggering kernel like in the standard Hawkes processes defined in \eqref{eq:lamba_multi}, one for each pair of event type combinations, and $x_i$ is the embedding of the $i$-th event obtained as the concatenation of the event type and temporal embeddings, $x_i = [e_{i}^{\rm enc} \  | \ t_i^{\rm enc}]$.
%
%
Note that $i$) we concatenate the embeddings unlike $x_i=t_i^{\rm enc}+e_i^{\rm enc}$ in \citet{pmlr-v119-zhang20q} to decouple the contribution of temporal and event type embeddings; $ii)$ in situations when longitudinal covariates are available, they can be readily incorporated as $x_i = [e_{i}^{\rm enc} \  | \ t_i^{\rm enc} \ | \ u_i^{\rm enc}]$, where $u_i^{\rm enc}$ is the representation for the covariates at time $t_i$ (see Experiments for an example and Appendix B for details); and $iii$) in \eqref{eq:h_q} we need not to normalize the weights as in \eqref{eq:att} because in \eqref{eq:h_q}, $h_j$ is meant to represent the aggregated, historical intensity rather than the average historical embedding.

Though for large values of $K$ it may seem inefficient to specify a triggering kernel for each combination, \eqref{eq:h_q} allows for the behavior of each event type combination to be modeled separately over time, unlike implicitly done so in attention-based approaches.
Below we will show that kernels for all event type combinations can be obtained efficiently by specifying a flexible kernel whose parameters are set and learned separately for each event type combination.

\begin{figure}[t!]
\centering
\includegraphics[width=0.9\columnwidth, height = 3.0cm]{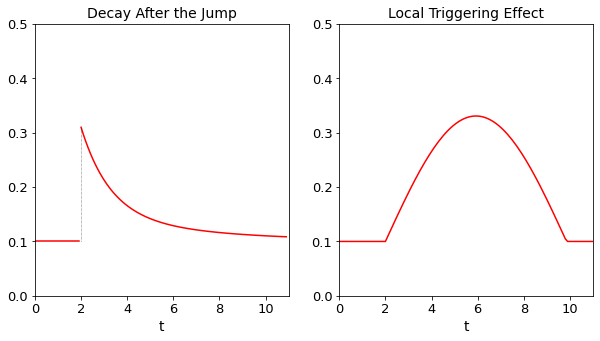}
\vspace{-4mm}
\caption{Comparison of a regular RQ kernel {\em versus} the proposed continuous local effect kernel, both triggered at $t=2$. It can be seen that the local effect is not a decreasing function with respect to the triggering time which cannot be captured with exponentiated absolute distance kernels.
\vspace{-2mm}
}
\label{fg:triggering_example}
\end{figure}

\subsection{Flexible Kernel with Decay and Gating Effects}
A natural choice for estimating a triggering function with a decaying effect over time is the squared exponential kernel, {\em i.e.}, the radial basis function (RBF), which is widely used in machine learning applications such as kernel machines \cite{708428} and Gaussian processes \citep{Rasmussen2009GaussianPF}.
Instead, we consider using the rational quadratic (RQ) kernel since it is a mixture of infinitely many squared exponential kernels, thus in principle, a more flexible yet easy to compute generalization of the squared exponential kernel \citep{Rasmussen2009GaussianPF}.
For a one dimensional process (with a single event type), the rational quadratic kernel is defined as follows
\begin{equation}\label{eq:rq}
    k(d) = \sigma^2 \left(1+\frac{d^2}{2\alpha \ell^2}\right)^{-\alpha} ,
\end{equation}
where $\alpha > 0$ and $\ell >0$ control the decay behavior, $\sigma$ is the scaling parameter, and $d$ is a pairwise distance function, {\em e.g.}, $d=|t_i-t_j|$ in \eqref{eq:h_q}.

\begin{figure}[t!]
\includegraphics[width=\columnwidth]{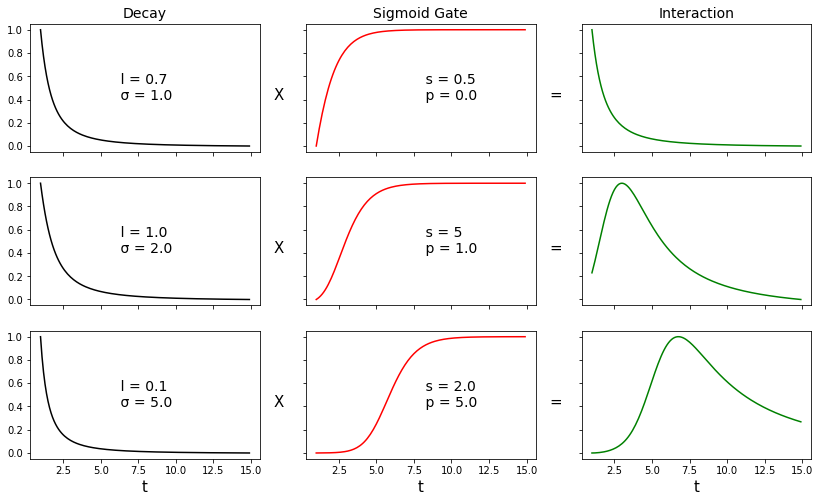}
\vspace{-7mm}
\caption{Examples of the interaction of the RQ kernel and the sigmoid gate with various parameters. The first row shows how we can still capture regular (monotonic) decay, while the second and the third rows show that we can also estimate local effects at various times, whose location is controlled by shift parameter $p$.}
\vspace{-4mm}

\label{fg:RQplusGateEx}
\end{figure}

The rational quadratic function is suitable to capture monotonically decaying effects, which are fairly standard in Hawkes processes.
However, underlying triggering kernels in real-world point processes can be often more complex.
As a particular example illustrated in Figure~\ref{fg:triggering_example}, consider a situation where the effect of a past event on the current event ($t=2$) does not decrease as the difference between event times grows but rather exhibits a localized behavior (at $t=6$) in which the effect is maximized at certain time difference (larger than zero).
Such local effect is not possible to capture with regular decaying functions. 
To capture these local effects, we multiply the rational quadratic kernel with a sigmoid gate function, thus allowing the resulting kernel to capture both decay and local triggering effects.
The proposed triggering kernel is then
\begin{equation}\label{eq:kernel}
    q(d;\theta) = \sigma^2\Bigg(1+\frac{d}{2\alpha\ell^2}\bigg)^{-\alpha} \bigg((1+e^{p-d})^{-s}\bigg) ,
\end{equation}
%
%
where the first term is the rational quadratic kernel defined in \eqref{eq:rq}, the second term is the generalized logistic function \cite{gupta2010generalized}, and $\theta=\{\sigma,\alpha,\ell,p,s\}$.
We modify the sigmoid function by adding the parameters $p>0$ and $s>0$, which denote the location and rate of the change, respectively.
As $p$ changes, the change-point moves across time, while s controls the spread of the change.
Moreover, $l$ and $\alpha$ still control the decay while $\sigma^2$ is now the scaling parameter of the whole function.
Figure~\ref{fg:RQplusGateEx} illustrates the behavior of \eqref{eq:kernel} for different parameters choices and how the kernel flexibly models different triggering functions.

\subsection{Efficient Learning of Event Type Kernel Pairs}
Provided the triggering kernel function being able to capture decay and local effects as defined above in \eqref{eq:kernel}, we can proceed to generalize it for the multivariate (multi-event) process.
Inconveniently, the number of parameters for the triggering kernels scales quadratically with the number of unique event types, $K$, which may render the modeling prohibitive even for moderately large $K$.
Specifically kernel parameters scale with ${\cal O}(PK^2)$,
%
%
%
where $P$ is the number of free parameters for the triggering kernel, {\em i.e.}, $|\theta|=5$ in the proposed approach according to \eqref{eq:kernel}.
For efficiency purposes, we estimate these parameters with feed-forward networks whose input are the concatenation of type embeddings as follows
%
\vspace{-3 mm}
\begin{equation}\label{eq:kernel_params}
    \theta_{v|u} = g(e_{v|u}^{\rm enc}) ,
\end{equation}
where $e_{v|u}^{\rm enc}=[e_{v}^{\rm enc} \ | \ e_{u}^{\rm enc}]$ is the concatenation of embeddings for event types $v$ and $u$, and $g(\cdot)$ is specified as a collection of individual fully connected networks via
\begin{align}\label{eq:ff_par}
    r_{v|u} & = {\rm softplus}(W_{r}^Te_{v|u}^{\rm enc} +b_r) ,
\end{align}
where $r_{v|u}$ is a component of $\theta$ and $r=\{\sigma,\alpha,\ell,p,s\}$.
The mapping in \eqref{eq:ff_par} indicates that each parameter in $\theta$ has a corresponding weight vector $W_r\in\mathbb{R}^{D}$ and bias $b_r\in\mathbb{R}$.
As a result, kernel parameters scale linearly with $P$ and $D$, {\em i.e.}, ${\cal O}(PD)$, thus effectively removing their dependency on $K$.
Further, the non-negativity of the parameters is ensured by the softplus activation function.
Note that one could also specify a single multilayer network for all parameters, however, in practice we found it more computational expensive but without substantial performance benefits.
%

\section{Predicting for Next Event in the Sequence}
The proposed model characterizes the underlying distribution of the point process by learning (temporal and event type) embeddings and the triggering kernels during the encoding stage.
This allows for added flexibility in the decoder architecture.
Specifically, by directly emphasizing on the prediction of next event (arrival) times and types during the decoding stage, we can encourage the predictive ability of the model.
Further, we introduce stochasticity into the decoder to increase the variation (uncertainty) in the predictions and thus being able to approximate the predictive conditional distribution function for the arrival times.

\subsection{Arrival Time Prediction}
Given the historical embeddings for each time step, we can focus on predicting the next arrival time at each step.
A straightforward way of obtaining arrival time predictions is to use a neural network whose input is the historical representation.
We define such a feed-forward neural network as follows
\begin{equation}\label{eq:dec}
    t^{\rm pred}_{j+1} = {\rm softplus}(W_t^T h_{j}+b_t) ,
\end{equation}
Where $h_j$ is the encoded history vector defined in \eqref{eq:h_q}, $W_{t}\in\mathbb{R}^D$ and $b_t\in\mathbb{R}$ are the weight and bias of the feed-forward network, respectively.
Note that $t^{\rm pred}_{j+1}$ stands for the arrival time prediction corresponding to event $j+1$, and that we use the information (history representation) we have up to event $j$ to predict the arrival time for the next time step $j+1$.
The softplus activation is added to the output layer to ensure the arrival time predictions stay non-negative.

\paragraph{Adding Stochasticity to Predictions}
%

As previously leveraged for generative models with adversarial learning \cite{NIPS2014_5ca3e9b1} \cite{DBLP:journals/corr/MirzaO14}, we introduce stochasticity into our arrival time prediction by sampling noise vectors from an easy to sample distribution and adding them to the history representations.
Specifically, the stochastic layer samples noise vectors of size $|h_j|$ from a uniform distribution.
Each noise vector is passed through a linear layer before being added to $h_j$, {\em i.e.},
\begin{equation}\label{eq:dec2}
    t_{j+1,m}^{\rm pred} = {\rm softplus}(W_t( W_h h_j + W_n n_{m})+b_t) ,
\end{equation}
where $n_m\sim U_{[0,1]}$ is a sample from a uniform distribution, and $W_h$ and $W_n$ are specified accordingly to \eqref{eq:dec}.
The collection of $M$ samples $\{t_{j+1,m}^{\rm pred}\}_{m=1}^M$ can be thought as implicitly sampled from the desired predicted conditional distribution, {\em i.e.}, $t_{j+1,m}^{\rm pred} \sim p(t_{j+1}^{pred}|h_j)$.
%

Since the loss function (defined below) takes the predictions from all of these samples and updates the parameters of the model based on all of them, we are in principle able to approximate the conditional distribution of the arrival times.
In practice, if distributional predictions are not a priority, we can simply summarize the $M$-sample empirical distributions with sample characteristics such as mean, median or standard deviation.
In the experiments, we use the sample mean as our final prediction for the arrival times.

\subsection{Event Type Prediction}
For multivariate event sequences, predicting the next event type is as important as predicting the arrival times.
Unlike the models that use conditional intensity values of each event type, we can specify another feed-forward network that uses the history vectors as input to predict the event types directly.
The event prediction is probabilistically defined as follows
\begin{equation}\label{eq:type}
    p(e^{\rm pred}_{j+1}|h_j) =  {\rm softmax}( W_{e}h_j +b_{e} ) ,
\end{equation}
where $W_{e}\in\mathbb{R}^{K\times D}$ and $b_{e}\in\mathbb{R}^K$ are the weights and the bias of the feed-forward network, $h_j$ is the historical representation of event $j$ defined in \eqref{eq:h_q}, and $e^{\rm pred}_{j+1} = {\rm argmax} \ p(e^{\rm pred}_{j+1}|h_j)\in(0,1)^{K}$, {\em i.e.}, the output of the softmax function produces event type probabilities that can be converted into predicted event types for time $j+1$.

\subsection{Loss Function}
%
We estimate the underlying distribution of the point process by learning the embeddings and triggering kernels, and use the next arrival times and event type as the target for the objective function.
In addition to emphasize predictive ability, this allows us to avoid the integral approximations traditionally used in intensity-based decoders. \cite{pmlr-v119-zhang20q,pmlr-v119-zuo20a,shchur2019intensity}.

Since the decoders in \eqref{eq:dec2} and \eqref{eq:type} predict both the arrival times and event types, we ought to specify two different loss functions to train the model. 
Specifically, we use the $\ell_1$ loss for the arrival times and the cross-entropy loss for the event types.
The complete loss is defined for the prediction of the next event at each observed timestamp as
\begin{equation*}
   L(S_n,\Phi) = \sum_j\bigg(\big|\tilde{t}^{\rm pred}_{j+1}-t_{j+1}\big|+
    \sum_{k=1}^K y_{k,j+1}\log(p_{k,j+1})\bigg) ,
\end{equation*}
where $\tilde{t}^{\rm pred}_{j+1}$ is a sample summary from \eqref{eq:dec2}, $p_{k,j+1}$ is the $k$-th element of $p(e^{\rm pred}_{j+1}=k|h_j)$ from \eqref{eq:type} and $y_{k,j+1}$ is the ground truth indicator of whether time $j+1$ is of event type $k$.
Moreover, $S_n$ indicates that the loss is for a single sequence (preventing notation overload), and $\Phi$ represents the set of all the parameters needed to be learned for the model.
Precisely, the parameters of $i$) the triggering kernel parameter estimator in \eqref{eq:kernel_params}; $ii$) the arrival time and event time predictors in \eqref{eq:dec2} and \eqref{eq:type}, respectively; $iii$) the scaling parameters of the temporal embeddings in \eqref{eq:time_emb}; and $iv$) the event type embeddings in \eqref{eq:type_emb}.
%
%
Finally, for the minimization problem we use the ADAM \cite{kingma2014adam}, provided it has proven to be computational efficient while requiring little parameter tuning compared to alternative algorithms.

\section{Experiments}
We present results on various synthetic and real-world datasets to compare the proposed {\em sigmoid gated Hawkes process} (SGHP) model against state-of-the-art Hawkes process models.
Source code in Pytorch is available at {\small\url{https://github.com/yamacisik/gated_kernel_tpp}}.

\subsection{Baseline Models}
We consider two attention-based Hawkes process models \cite{pmlr-v119-zhang20q,pmlr-v119-zuo20a} and an RNN-based log-normal mixture decoder model \cite{shchur2019intensity}.
Further, we introduce a fourth baseline, namely, an attention-based log-normal mixture model, by replacing the RNN encoder in the original log-normal model with an attention encoder.
We briefly describe the baseline models below.

\textbf{Self Attentive Hawkes Process (SAHP):}
An attention-based model that uses multiple layers of multi-headed attention to model the intensity function based on historical embeddings.
\citet{pmlr-v119-zhang20q} introduced time-shifted positional embeddings to inject information on event time differences in addition to event temporal position.
They learn the intensity function of the event sequence via maximum likelihood while using Monte Carlo sampling to approximate the cumulative intensity function.

\textbf{Transformer Hawkes Process (THP):}
The model by \citet{pmlr-v119-zuo20a} is a variant of an attention-based approach similar to SAHP.
In addition to model the log-likelihood of the intensity function, they incorporate event time and type prediction losses for the training of their model.
Like the SAHP model, the intensity function is a non-linear transformation of the base intensity with additive decay, thus the triggering kernels are not directly modeled.

\textbf{Log-Normal Mixture Model (LMM):}
The log-normal mixture model \cite{shchur2019intensity} is an intensity-free approach that models the conditional probability distribution of arrival time with a log-normal mixture model.
The model uses an RNN to encode the history representations and the log-likelihood of the conditional event probabilities as their loss function, thus avoiding Monte Carlo sampling.
However, to obtain the intensity function, they still need to approximate the integral of the cumulative distribution function.

\textbf{LMM with Attention (LMMA):}
We replace the RNN decoding of the log-normal mixture model with an attention mechanism proposed in the SAHP model.
This simple extension of LMM is introduced here to understand the performance of attention models with an intensity-free formulation.
This is conceptually similar to our approach in that we summarize histories leveraging ideas from attention, but unlike other attention models, arrival event times and event types are predicted without the need to directly estimate the cumulative intensity function.

\subsection{Datasets}
We use a synthetic dataset sampled from a 2D Hawkes process and three real-word datasets previously used by the baseline models.
Further, we consider a COVID-19 inpatient dataset collected at our Institution.
Table 7 in the Appendix presents summary statistics for these datasets.
With the exception of the COVID-19 dataset, all real-world datasets are publicly available ({\small\url{https://drive.google.com/drive/folders/1OB1Mcns6qvnkZ48MCnZbtvB9Rp5Dtfp6}}).

\textbf{2D Hawkes (HP):}
We use the same 2D Hawkes dataset used  by in the SAHP \cite{pmlr-v119-zhang20q}.
For reproducibility, we resampled the dataset from the tick library with a fixed random seed.
Appendix A provides details on the triggering kernels and the baseline intensities used for the synthetic dataset.

\textbf{Stackoverflow (SOF):}
The data set consists of users of the famous question-and-answer website Stack Overflow.
Users get medals based on their questions or answers, {\em e.g.}, good question, good answer, {\em etc}.
Each user is modeled as a sequence with event types representing medals obtained over time.

\textbf{Mimic-II (MMC):}
This electronic healthcare records dataset tracks the journey of patients through their hospital intensive care unit (ICU) stay over a period of seven years.
Each patient sequence consists of timestamps and diagnoses for each visit.

\textbf{Retweets (RT):}
The dataset tracks re-tweets made on particular tweets on the social media website Twitter.
Every time a tweet is re-tweeted, the popularity of the re-tweeter and the timestamp of the retweet are recorded.
There are three different re-tweeter categories based on their number of followers.
Due to the wide range of the inter-arrival times, arrival time results are made relative by scaling over the maximum inter-arrival time.

\begin{figure}[t!]
\includegraphics[width=0.96\columnwidth,height = 8.3cm]{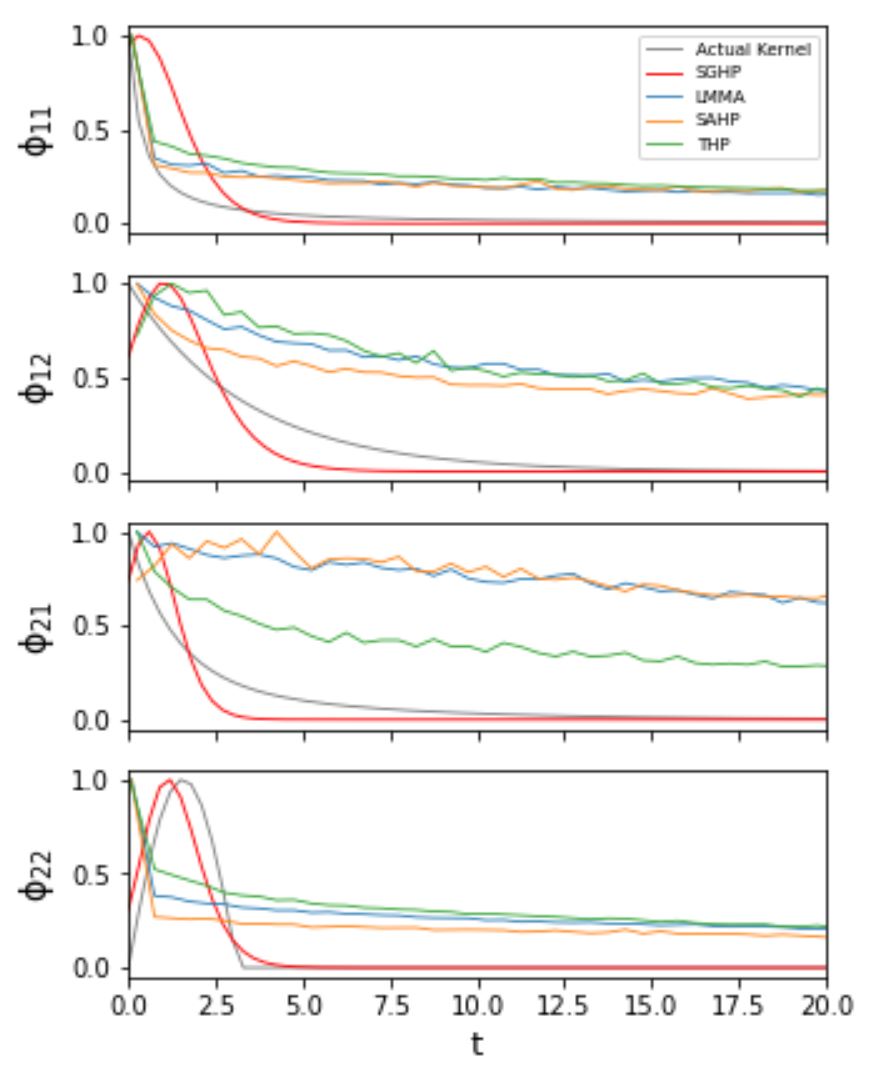}
\vspace{-6mm}
\caption{Comparing triggering kernel predictions across models. The x-axis denotes time and the y-axis represents the kernel values. We see that the sigmoid-gated kernel model is able to learn both the exponential kernel and sinusoidal kernel more accurately.}
\vspace{-6mm}
\label{fg:trigger_est}
\end{figure}

\textbf{COVID-19 Emergency Department (COVID):}
This dataset tracks the journey of COVID-19 patients admitted for patient care at [Name] Hospital Emergency Department (ED) between January of 2020 and December of 2021.
We consider five different event types, namely, admission to ED for patient care, admission to ICU, admission to a step-down unit, discharge, and death.
Further, the dataset also includes vitals measured between events.
Specifically, five vitals: pulse oximetry (SpO2), mean arterial pressure (MAP), blood pressure (BP), temperature, and respiratory rate (RR).
%
\section{Results}
\subsection{Capturing Triggering Kernels}
We present the estimated kernels for all the models against the ground-truth triggering functions for the synthetic data and compare the goodness-of-fit to the underlying distribution.
Since our approach learns the parameters for each triggering kernel, we can plot the estimation directly as a function of time.
For the attention-based models, we use the attention scores for each time difference observed in the data as proxy.
We further split these scores across event type pairings.
Moreover, to reduce the fluctuations in the attention-based scores, we pass them through a moving average filter where we take mean scores over a small time window ($t = 0.5$).
Figure~\ref{fg:trigger_est} shows the estimates for all models.
As expected, the proposed gated kernel (SGHP) is the best match for all of the underlying triggering kernel pairings.
Importantly, unlike the other models, it can accommodate to different event type pairings and capture both decaying and local effects.


\begin{table}[t!]
\caption{RMSE for last event arrival time prediction.} \label{F-1}
\vspace{2mm}
\begin{tabular}{l|rrrrr}
\textbf{Dataset}&LMM&LMMA&THP&SAHP&SGHP \\
\hline
HP &  2.25  &  2.23  &  2.42 &  2.29  & \textbf{2.11} \\ 
MMC & 0.88 &  0.80  &  0.859  &  0.821  & \textbf{0.79}  \\ 
SOF &1.46&1.49&1.66&1.55&\textbf{1.30} \\ 
RT &0.61&0.64&0.18&0.12&\textbf{0.08} \\ 
COVID & 341.5&356.1&374.1&383.2& \textbf{332.2} \\
\end{tabular}
\label{tb:rmse}

\end{table}

\subsection{Sequence Prediction}
Predicting the next event arrival time is a key objective when modeling event sequences.
To capture the ability all models have to learn from the sequential event information, we report the event type and event time prediction results for the last observed event in each sequence.
Specifically, we report (micro) F1 scores for event types, except for the COVID data for which a high class imbalance is observed, {\em i.e.}, $4\%$ mortality.
Instead, results for mortality prediction are provided below using average precision score as performance metric.
For event arrival times, we report the mean squared error (RMSE).
Tables \ref{tb:rmse} and \ref{tb:f1} show RMSE and F1 scores, respectively, for the last event predictions in the test sets of all datasets and models being evaluated.
In terms of RMSE, the proposed SGHP consistently outperforms the other approaches, while in terms of event type prediction, SGHP outperforms the others on the HP and RT datasets and delivers competitive results on MMC and SOF.
Interestingly the best performing event type prediction model in MMC is LMMA, which is also introduced in this paper.  

\begin{table}[t!]
\caption{Micro F1 score for last event type prediction.}
\vspace{2mm}
\begin{tabular}{l|rrrrr}
\textbf{Dataset}&LMM&LMMA&THP&SAHP&SGHP \\
\hline 
HP & 0.573 & 0.550 & 0.575 & 0.585 & \textbf{0.610} \\ 
MMC&0.892&\textbf{0.900}&0.877&0.646&0.810 \\ 
SOF&0.384&0.375&\textbf{0.411}&0.242&0.389 \\ 
RT&0.544&0.528&0.539&0.531 &\textbf{0.605} \\
\end{tabular}
\label{tb:f1}
\end{table}

\subsection{Mortality Predictions for Covid-19 Patients}
Complementing the real-world datasets in Table \ref{tb:f1}, we also seek to test the proposed model in a more realistic healthcare setting.
Specifically, we compare models in terms of their ability to predict patient mortality.
We treat the mortality prediction as a binary problem, {\em e.g.}, death {\em vs.} survival.
Since there is a significant class imbalance, we report the average precision score (APS) as opposed to F1 or area under the receiving operating characteristic (see Appendix for these).
Moreover, we incorporate patient vitals as longitudinal covariates as a means to showcase that the model can readily process covariate information.
Existing approaches do not consider covariates, but we have incorporated them as embeddings as described in Appendix B for fair comparison.
Table~\ref{tb:aps} shows mortality prediction results on both variations of the COVID data (with and without vitals).Without covariates, SGHP delivers average performance while it outperforms the other approaches when longitudinal covariates (vitals) are included.

\vspace{-3mm}
\begin{table}[h]
\caption{APS results for mortality prediction on COVID data with and without patient vitals (COVID+v).}
\vspace{2mm}
\begin{tabular}{l|rrrrr}
\textbf{Dataset}&LMM&LMMA&THP&SAHP&SGHP\\
\hline 
COVID & \textbf{0.18}&0.16& 0.07&0.004&0.12 \\
COVID+v & 0.12&0.21& 0.1&0.005&\textbf{0.22} \\
\end{tabular}

\label{tb:aps}
\end{table}
\subsection{Model Complexity}
RNN-based and attention-based encoders tend to suffer from high complexity and memory requirements due to the large number of parameters that need to be estimated.
Table \ref{tb:count} reports the total number of estimated parameters (in thousands) for each model across all the data sets.
Note that all datasets use the same model architectures.
We see that our model often requires a significantly smaller number of parameters for all prediction tasks, while delivering comparable or superior performance on both arrival times and event type prediction.

\begin{table}[t!]
\caption{Total number of learned parameters (in thousands).}
\vspace{2mm}
\begin{tabular}{l|rrrrr}
\textbf{Dataset}&LMM&LMMA&THP&SAHP& SGHP \\
\hline 
HP & 17.2 & ~17.2 & 25.3& 4.1 & \textbf{2.7} \\ 
MMC&17.8&19.6&189&6.5&\textbf{4.2} \\ 
SOF&15.7&17.8&1595&4.8&\textbf{3.7} \\ 
RT&14.8&17.2&41.7&4.2 &\textbf{2.4} \\   
COVID&14.9&17.3&23.7&4.2 &\textbf{3.8} \\
\hline
Average & 16.9 & 17.8 & 417.8 & 4.8& \textbf{3.1}
\end{tabular}
\label{tb:count}
\end{table}

\subsection{Kernels Interpretation for COVID Data}
\begin{figure}[h!]
\includegraphics[width=\columnwidth]{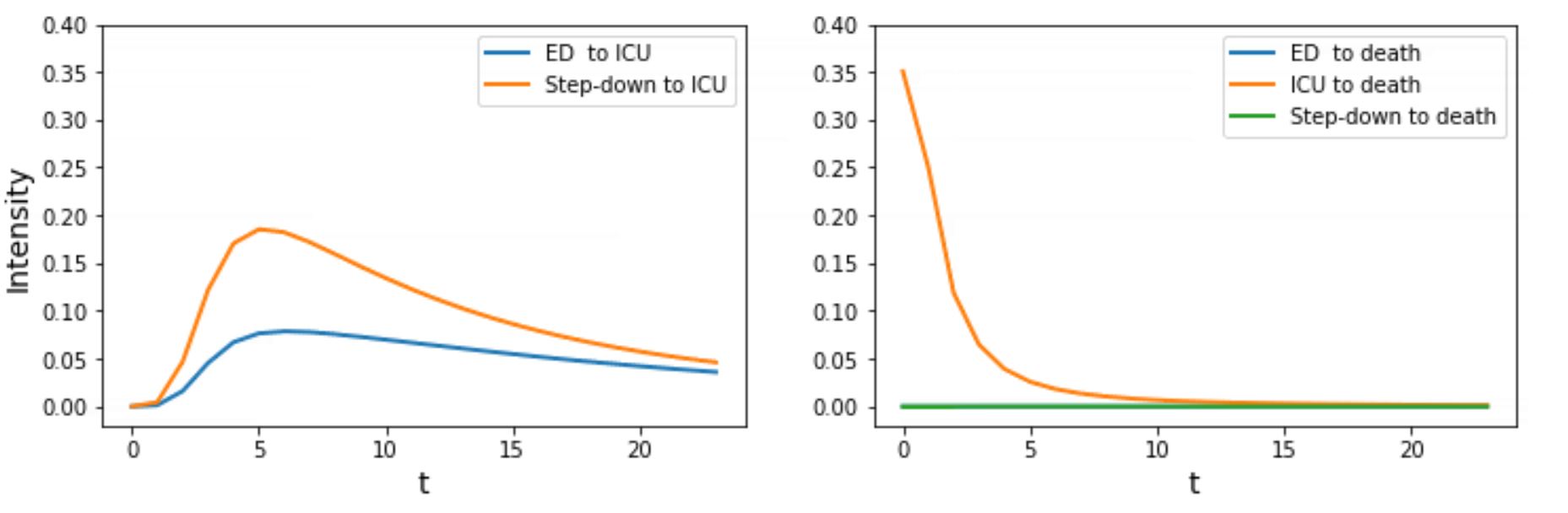}
\includegraphics[width=\columnwidth]{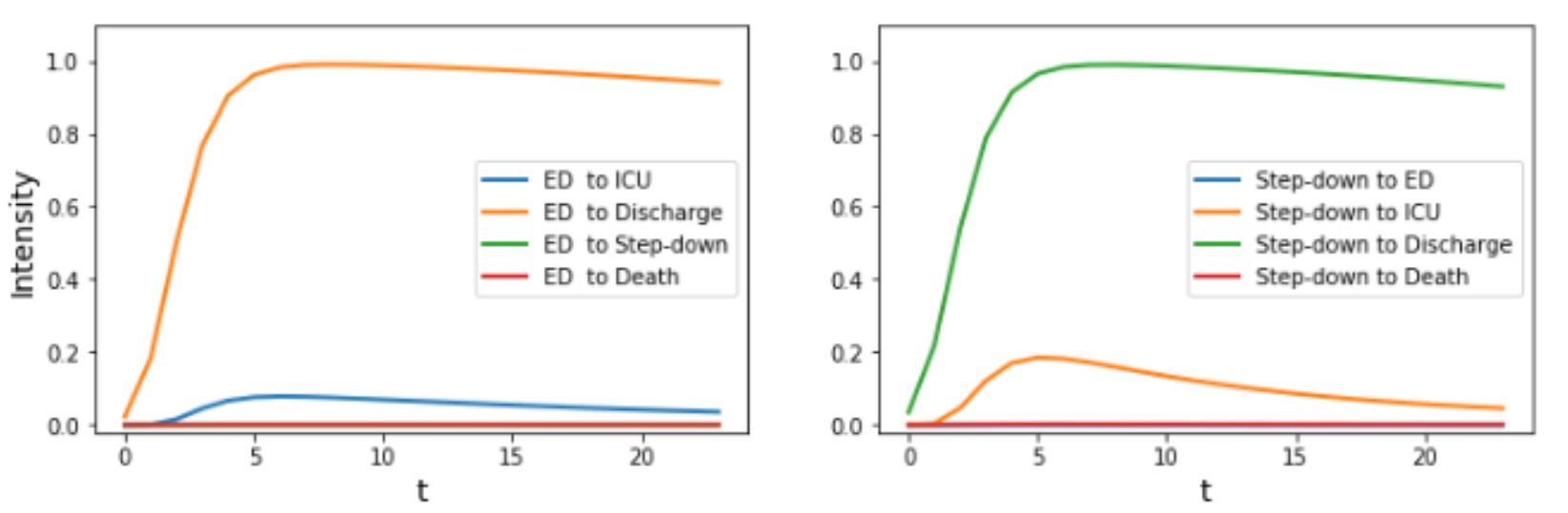}
\vspace{-8mm}
\caption{Estimated SGHP triggering kernels for various combinations of event type pairs for the COVID-19 data.}
\vspace{-3mm}
\label{fg:icu_mort}
\end{figure}

Our model learns a triggering kernel for each event type pairing.
This allows one to interpret the results in terms of intensity functions.
Specifically, we report the likelihood of being admitted to ICU and mortality after observing that a patient is admitted to the ED, ICU, or a step-down unit.
Figure~\ref{fg:icu_mort} shows the learned triggering kernels for some of these possible transitions.
As expected, the likelihood of mortality is by far sharper for patients already admitted to the ICU, relative to patients in ED beds or step-down units.
Moreover, the likelihood of being admitted to the ICU is much higher for patients originating from a step-down unit.

\section{Conclusion}
We introduced a flexible, efficient, and interpretable approach for modeling Hawkes processes with an encoder-decoder structure.
Our approach replaced the complex and more difficult to interpret attention mechanism with directly learning the underlying triggering kernels of the point processes.
The proposed encoder uniquely combines a regular decaying kernel with a sigmoid gate, thus allowing it to capture a wide range of triggering functions otherwise not possible with other attention-based approaches.
We further generalized our method for the multivariate Hawkes process and incorporated longitudinal covariates into the temporal process characterization.
Finally, the experiments demonstrated that the proposed approach is superior to state-of-art methods in terms of predictive and computational performance, interpretation of the kernels learned from data.

%
\bibliography{main}

\begin{thebibliography}{27}
\providecommand{\natexlab}[1]{#1}
\providecommand{\url}[1]{\texttt{#1}}
\expandafter\ifx\csname urlstyle\endcsname\relax
  \providecommand{\doi}[1]{doi: #1}\else
  \providecommand{\doi}{doi: \begingroup \urlstyle{rm}\Url}\fi

\bibitem[Alvari \& Shakarian(2019)Alvari and Shakarian]{8855306}
Alvari, H. and Shakarian, P.
\newblock Hawkes process for understanding the influence of pathogenic social
  media accounts.
\newblock In \emph{International Conference on Data Intelligence and Security},
  2019.

\bibitem[Bacry et~al.(2015)Bacry, Mastromatteo, and Muzy]{bacry2015hawkes}
Bacry, E., Mastromatteo, I., and Muzy, J.-F.
\newblock Hawkes processes in finance.
\newblock \emph{Market Microstructure and Liquidity}, 2015.

\bibitem[Bao et~al.(2017)Bao, Kuang, Peissig, Page, and
  Willett]{pmlr-v68-bao17a}
Bao, Y., Kuang, Z., Peissig, P., Page, D., and Willett, R.
\newblock Hawkes process modeling of adverse drug reactions with longitudinal
  observational data.
\newblock In \emph{Machine Learning for Healthcare Conference}, 2017.

\bibitem[Chen \& Hall(2016)Chen and Hall]{chen2016nonparametric}
Chen, F. and Hall, P.
\newblock Nonparametric estimation for self-exciting point processes—a
  parsimonious approach.
\newblock \emph{Journal of Computational and Graphical Statistics}, 2016.

\bibitem[Chiang et~al.(2021)Chiang, Liu, and Mohler]{CHIANG2021}
Chiang, W.-H., Liu, X., and Mohler, G.
\newblock Hawkes process modeling of covid-19 with mobility leading indicators
  and spatial covariates.
\newblock \emph{International Journal of Forecasting}, 2021.

\bibitem[Cox \& Isham(1980)Cox and Isham]{cox1980point}
Cox, D. and Isham, V.
\newblock \emph{Point Processes}.
\newblock Chapman \& Hall/CRC Monographs on Statistics \& Applied Probability.
  Taylor \& Francis, 1980.

\bibitem[Du et~al.(2016)Du, Dai, Trivedi, Upadhyay, Gomez-Rodriguez, and
  Song]{du2016recurrent}
Du, N., Dai, H., Trivedi, R., Upadhyay, U., Gomez-Rodriguez, M., and Song, L.
\newblock Recurrent marked temporal point processes: Embedding event history to
  vector.
\newblock In \emph{International Conference on Knowledge Discovery and Data
  Mining}, 2016.

\bibitem[Goodfellow et~al.(2014)Goodfellow, Pouget-Abadie, Mirza, Xu,
  Warde-Farley, Ozair, Courville, and Bengio]{NIPS2014_5ca3e9b1}
Goodfellow, I., Pouget-Abadie, J., Mirza, M., Xu, B., Warde-Farley, D., Ozair,
  S., Courville, A., and Bengio, Y.
\newblock Generative adversarial nets.
\newblock In Ghahramani, Z., Welling, M., Cortes, C., Lawrence, N., and
  Weinberger, K.~Q. (eds.), \emph{Advances in Neural Information Processing
  Systems}, 2014.

\bibitem[Gupta \& Kundu(2010)Gupta and Kundu]{gupta2010generalized}
Gupta, R.~D. and Kundu, D.
\newblock Generalized logistic distributions.
\newblock \emph{Journal of Applied Statistical Science}, 2010.

\bibitem[Hawkes(1971)]{10.2307/2334319}
Hawkes, A.~G.
\newblock Spectra of some self-exciting and mutually exciting.
\newblock \emph{Biometrika}, 1971.

\bibitem[Hearst et~al.(1998)Hearst, Dumais, Osuna, Platt, and
  Scholkopf]{708428}
Hearst, M., Dumais, S., Osuna, E., Platt, J., and Scholkopf, B.
\newblock Support vector machines.
\newblock \emph{IEEE Intelligent Systems and their Applications}, 1998.

\bibitem[Kingma \& Ba(2015)Kingma and Ba]{kingma2014adam}
Kingma, D.~P. and Ba, J.
\newblock Adam: A method for stochastic optimization.
\newblock In \emph{International Conference on Learning Representations}, 2015.

\bibitem[Lee \& Seo(2017)Lee and Seo]{LEE2017174}
Lee, K. and Seo, B.~K.
\newblock Marked hawkes process modeling of price dynamics and volatility
  estimation.
\newblock \emph{Journal of Empirical Finance}, 2017.

\bibitem[Lewis \& Mohler(2011)Lewis and Mohler]{lewis2011nonparametric}
Lewis, E. and Mohler, G.
\newblock A nonparametric em algorithm for multiscale hawkes processes.
\newblock \emph{Journal of Nonparametric Statistics}, 2011.

\bibitem[Mei \& Eisner(2017)Mei and Eisner]{mei2016neural}
Mei, H. and Eisner, J.~M.
\newblock The neural hawkes process: A neurally self-modulating multivariate
  point process.
\newblock In \emph{Advances in Neural Information Processing Systems}, 2017.

\bibitem[Mirza \& Osindero(2014)Mirza and
  Osindero]{DBLP:journals/corr/MirzaO14}
Mirza, M. and Osindero, S.
\newblock Conditional generative adversarial nets.
\newblock \emph{CoRR}, 2014.

\bibitem[Ozaki(1979)]{ozaki1979maximum}
Ozaki, T.
\newblock Maximum likelihood estimation of hawkes' self-exciting point
  processes.
\newblock \emph{Annals of the Institute of Statistical Mathematics}, 1979.

\bibitem[Pascanu et~al.(2013)Pascanu, Mikolov, and Bengio]{pmlr-v28-pascanu13}
Pascanu, R., Mikolov, T., and Bengio, Y.
\newblock On the difficulty of training recurrent neural networks.
\newblock In \emph{International Conference on Machine Learning}, 2013.

\bibitem[Rasmussen \& Williams(2005)Rasmussen and
  Williams]{Rasmussen2009GaussianPF}
Rasmussen, C.~E. and Williams, C. K.~I.
\newblock \emph{Gaussian Processes for Machine Learning (Adaptive Computation
  and Machine Learning)}.
\newblock The MIT Press, 2005.

\bibitem[Rizoiu et~al.(2017)Rizoiu, Lee, Mishra, and Xie]{rizoiu2017tutorial}
Rizoiu, M.-A., Lee, Y., Mishra, S., and Xie, L.
\newblock A tutorial on hawkes processes for events in social media.
\newblock \emph{arXiv preprint arXiv:1708.06401}, 2017.

\bibitem[Shchur et~al.(2020)Shchur, Biloš, and
  Günnemann]{shchur2019intensity}
Shchur, O., Biloš, M., and Günnemann, S.
\newblock Intensity-free learning of temporal point processes.
\newblock In \emph{International Conference on Learning Representations}, 2020.

\bibitem[Steinruecken et~al.(2019)Steinruecken, Smith, Janz, Lloyd, and
  Ghahramani]{Steinruecken2019}
Steinruecken, C., Smith, E., Janz, D., Lloyd, J., and Ghahramani, Z.
\newblock \emph{The Automatic Statistician}.
\newblock 2019.

\bibitem[Sun et~al.(2021)Sun, Sun, Dong, Shi, and Huang]{9502579}
Sun, Z., Sun, Z., Dong, W., Shi, J., and Huang, Z.
\newblock Towards predictive analysis on disease progression: A variational
  hawkes process model.
\newblock \emph{IEEE Journal of Biomedical and Health Informatics}, 2021.

\bibitem[Vaswani et~al.(2017)Vaswani, Shazeer, Parmar, Uszkoreit, Jones, Gomez,
  Kaiser, and Polosukhin]{46201}
Vaswani, A., Shazeer, N., Parmar, N., Uszkoreit, J., Jones, L., Gomez, A.~N.,
  Kaiser, {\L}., and Polosukhin, I.
\newblock Attention is all you need.
\newblock In \emph{Advances in Neural Information Processing Systems}, 2017.

\bibitem[Yang et~al.(2018)Yang, Liu, Chen, and
  Hawkes]{doi:10.1080/14697688.2017.1403156}
Yang, S.~Y., Liu, A., Chen, J., and Hawkes, A.
\newblock Applications of a multivariate hawkes process to joint modeling of
  sentiment and market return events.
\newblock \emph{Quantitative Finance}, 2018.

\bibitem[Zhang et~al.(2020)Zhang, Lipani, Kirnap, and
  Yilmaz]{pmlr-v119-zhang20q}
Zhang, Q., Lipani, A., Kirnap, O., and Yilmaz, E.
\newblock Self-attentive {H}awkes process.
\newblock In \emph{International Conference on Machine Learning}, 2020.

\bibitem[Zuo et~al.(2020)Zuo, Jiang, Li, Zhao, and Zha]{pmlr-v119-zuo20a}
Zuo, S., Jiang, H., Li, Z., Zhao, T., and Zha, H.
\newblock Transformer {H}awkes process.
\newblock In \emph{International Conference on Machine Learning}, 2020.

\end{thebibliography}
\bibliographystyle{icml2022}

\newpage
\appendix
\twocolumn

\section{Synthetic Data Set}
\begin{figure}[h]
\includegraphics[width=\columnwidth]{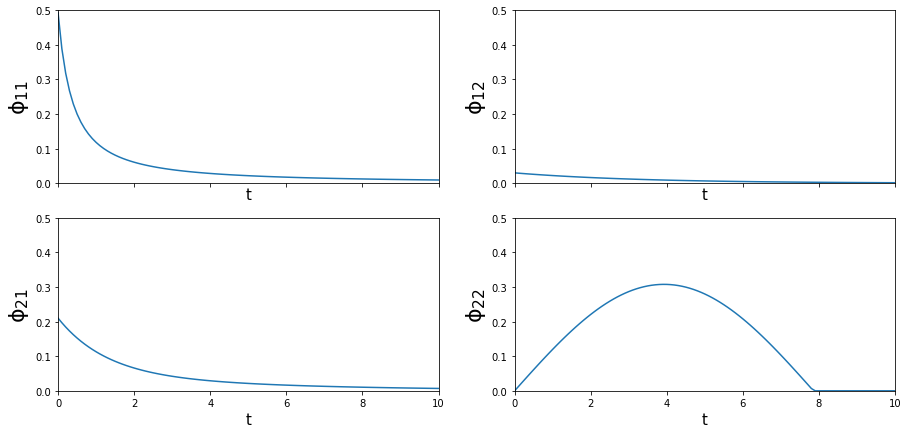}
\caption{The triggering functions for all event type combinations of the synthetic dataset with their formulas.}
\vspace{-4mm}
\label{fg:synth}
\end{figure}

Below we provide the formulation for each triggering function in the synthetic dataset.
The baseline intensities are $\mu_1 = 0.1$ and $\mu_2 = 0.2$.
There are three regular decaying kernels and a sinusoidal kernel with a local effect.
Figure \ref{fg:synth} illustrates each triggering kernel.
\begin{align}
    \begin{aligned}
    \phi_{11}(t) & = 0.2 \times  t (0.5 +t)^{-1.3} \\
    \phi_{12}(t) & = 0.03 \times   e^{-0.3t}  \\
%
     \phi_{12}(t) & = 0.05 \times   e^{-0.2t}+  0.16 \times  e^{-0.8t} \\
         \phi_{22}(t) & = \max\big(0,\sin(t)/8)\big) \text{       for} \ 0\leq t \leq 4 
    \end{aligned}
\end{align}

\section{Incorporating Longitudinal Covariates}

%
We generalize the input vector to include the addition of longitudinal covariates.
%
In order to inject the information from the covariates, we create a linear transformation that outputs a covariate embedding $u_i^{\rm enc}$ of dimensionality $D$; consistent with the dimension of the temporal and event type embeddings.
We modify the concatenated embedding vector as follows
\begin{equation}
    x_i = [ e_{i}^{\rm enc} \ | \ t_{i}^{\rm enc} \ | \ u_i^{\rm enc} ] ,
\end{equation}
where $u_i^{\rm enc}$ follows from
\begin{equation}
    u_i^{\rm enc} = W_u z_i + b_u ,
\end{equation}
Where $z_i$ is the vector of covariates for and event at time $t_i$, $W_u$ and $b_u$ are weights and the bias of the linear layer, respectively.
This structure can be generalized to all encoder based models, and is used for the COVID-19 prediction results with patient vitals.
\vspace{2cm}
\section{Additional Results}
Tables \ref{tb:covid_auc} and \ref{tb:covid_f1} present additional results for event type and mortality predictions on the COVID-19 dataset, namely  area under the receiving operating characteristic (AUROC)
and F1 scores, respectively. In addition Figure 6 shows the ROC and Precision-Recall Curves for the same task.
\begin{table}[!hp]
\caption{AUROC results for mortality prediction on COVID data with and without patient vitals (COVID+v).}
\vspace{2mm}
\begin{tabular}{l|rrrrr}
\textbf{Dataset}&LMM&LMMA&THP&SAHP&SGHP\\
\hline 
COVID & 0.752&\textbf{0.801}& 0.450&0.448&0.800 \\
COVID+v & 0.790&\textbf{0.880}& 0.745&0.482&0.840 \\
\end{tabular}
\label{tb:covid_auc}
\end{table}

\begin{table}[!hp]
\caption{F1 results for last event type prediction for the COVID data with and without patient vitals (COVID+v).}
\vspace{2mm}
\begin{tabular}{l|rrrrr}
\textbf{Dataset}&LMM&LMMA&THP&SAHP&SGHP\\
\hline 
COVID & 0.901&0.888& \textbf{0.932}&0.762&0.888 \\
COVID+v & 0.886&0.901& \textbf{0.921}&0.793& 0.893 \\
\end{tabular}
\label{tb:covid_f1}
\end{table}

\newpage
\onecolumn

\begin{figure*}[h]
\begin{center}
\includegraphics[scale = 0.65]{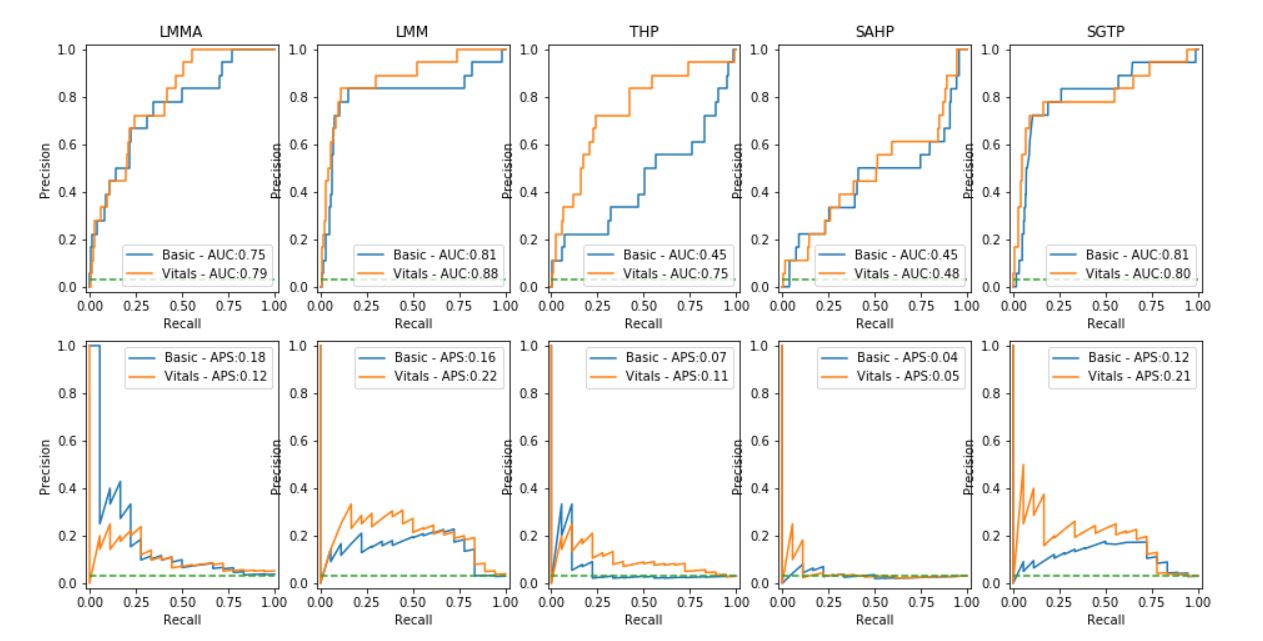}
\caption{ROC and Precision-Recall Curves for Mortality Prediction.}
\end{center}
\end{figure*}

\section{Dataset Statistics} 

Table 7 presents additional information on the data sets, such as the number of unique events, mean sequence lengths, the total number of sequences, and covariates.

\begin{table}[!hp]
\caption{Summary statistics of the datasets used in the experiments.} 
\label{RMSE}
\scalebox{1.0}{
\begin{tabular}{l|rrrr}
\hline
Dataset & Number o Types & Mean Seq. Length  & Number of Sequences & Number of Covariates \\ 
HP& 2 &150 &4000 &-\\
 MMC& 75& 4 & 650 &-\\
 SOF & 22 &250 &6633&-\\
 RT& 3 &109 &24000 &-\\
 COVID & 5 &4 & 4112 & -\\
 COVID+v  & 5 &4 & 4112 &5\\

\end{tabular}}

\end{table}


\end{document}